\documentclass[]{fairmeta}

\usepackage{xspace}
\usepackage{xcolor}
\newcommand{\name}{\textsc{Gamut}\xspace}
\newcommand{\metric}{\name score\xspace}
\newcommand{\judge}{Gemini~3.1~Pro\xspace}
\newcommand{\nq}{1{,}813\xspace}
\newcommand{\bestmm}{58.7\%\xspace}
\newcommand{\bestmeets}{55\%\xspace}
\newcommand{\avgchecks}{15.2\xspace}
\newcommand{\accheck}{5.9\xspace}
\newcommand{\valcheck}{5.8\xspace}
\newcommand{\ctxcheck}{3.5\xspace}
\usepackage{fvextra}
\newcommand{\promptbox}[2]{%
  \begin{tcolorbox}[breakable, colback=gray!4, colframe=gray!55, boxrule=0.4pt,
    fonttitle=\bfseries, title={#1}, left=2pt, right=2pt, top=2pt, bottom=2pt]
    \VerbatimInput[breaklines=true, breakanywhere=true,
      breaksymbolleft={\textcolor{gray!60}{\tiny$\hookrightarrow$}},
      fontsize=\scriptsize]{#2}%
  \end{tcolorbox}%
}

\title{Two-Level Meta-Rubrics for Evaluating Open-Ended Generation: GAMUT, a Benchmark for Factual Completeness}

\author[1]{Xilun Chen}
\author[1]{Zhaleh Feizollahi}
\author[1]{Ross Goodwin}
\author[1]{Seungwhan Moon}
\author[1]{Scott Yih}
\author[1]{Pinar Donmez}
\author[1]{Babak Damavandi}
\author[1]{Luna Dong}

\affiliation[1]{Meta AI}

\abstract{%
Rubric-based evaluation of open-ended generation faces a fundamental tension between \emph{expressiveness} and \emph{reliability}.
Authoring a faithful rubric requires expressing the structure of the space of good answers: open-ended sets of acceptable options, ordered processes, and the relative importance of facts.
Grading with the rubric requires a judge to score consistently, and judges are far more reliable on \emph{flat}, binary checks than on rich structure.
We resolve this tension with a \textbf{two-level meta-rubric framework}.
A structured meta-rubric captures the grading criteria at authoring time, and fixed mechanical rules compile it into a flat checklist of binary, machine-gradable checks that an LLM judge scores reliably at evaluation time.
We instantiate the framework as \name (\emph{Grounded Assessment of Multimodal Factuality}), a benchmark for factual completeness in long-form generation.
\name comprises \nq questions grounded in real wearable imagery across 10 diverse domains, each paired with an evidence-backed rubric verified by expert human annotators.
Evaluating 14 frontier and open-weight models, we find \name genuinely challenging (best score \bestmm from Gemini 3.1 Pro), highly discriminative, and robust to the choice of judge.

}

\date{\today}
\correspondence{Xilun Chen at \email{xilun@meta.com}}

\metadata[Data]{\url{https://huggingface.co/datasets/facebook/GAMUT}}
\metadata[Code]{\url{https://github.com/facebookresearch/GAMUT}}

\begin{document}

\maketitle

\begin{figure*}[t]
  \centering
  \includegraphics[width=\textwidth]{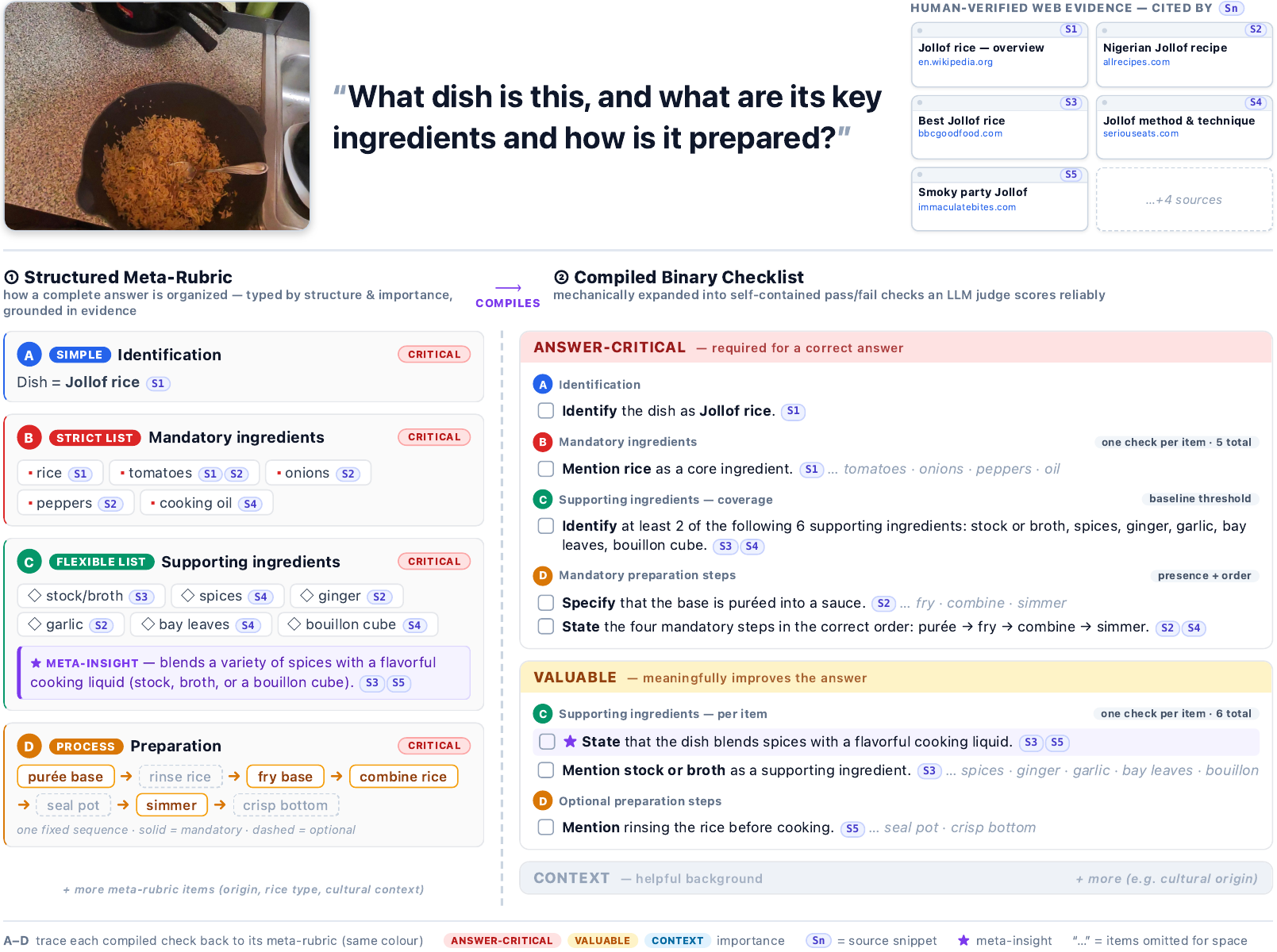}
  \caption{A representative example (jollof rice) from \name (some rubrics not shown for brevity).
  Top: an \emph{everyday deep-research} question and human-verified web evidence used to ground the rubrics.
  Left: the structured meta-rubric, with typed components (Simple Knowledge, Strict List, Flexible List with a meta-insight, Process), importance tiers, and citations.
  Right: the compiled binary checklist organized by importance tier, showing how each meta-rubric item mechanically expands into self-contained pass/fail checks.
  More details can be found in Section~\ref{section:framework}, and complete examples are given in Appendix~\ref{appendix:examples}.
  }
  \label{fig:teaser}
\end{figure*}

\section{Introduction}
\label{section:intro}

Most work on evaluating the factuality of long-form generation has focused on \emph{precision}: of the claims a response makes, what fraction are correct.
The dominant decompose-search-verify (DSV) paradigm breaks a response into atomic claims and verifies each against retrieved web documents, as in FActScore~\citep{factscore} and VeriScore~\citep{veriscore}.
Comparatively little attention has gone to \emph{recall}: whether a response contains all the information a complete answer should, which a few recent works target directly~\citep{verifact}.
But both lines rest on a more questionable assumption: that the factual quality of an answer is captured by a list of independent boolean fact-checks.
Reducing factuality to such a checklist is convenient for scoring, but it misrepresents how the quality of a knowledge-seeking answer is determined.

Several common situations break this model.
Many questions are open-ended, with a large or unbounded set of acceptable facts: asked which traditional dishes use a given ingredient, a good answer names enough of them, but no single dish is required and two answers listing different valid dishes can be equally good, which a fixed list of independent checks cannot express.
Other questions turn on a process whose order matters, such as how an everyday object reached its modern form, where stating the milestones out of sequence is an error even when each is individually correct.
Treating every fact as an independent, individually required checkbox discards these distinctions, along with the ability to ask whether an answer covers enough of an open set or respects a necessary order.
Measuring factual completeness faithfully therefore calls for a richer representation: a \emph{structured meta-rubric} of the facts a complete answer should contain, organized by importance and by the relations, orderings, and groupings among them.

Such structure, however, is not suited to reliable automatic evaluation, since judges are far more consistent scoring flat, binary, independently checkable items than interpreting rich structure.
\name resolves this tension with a \textbf{two-level representation}: a structured meta-rubric captures the organization and importance of the required content, which is then mechanically compiled into a flat checklist of binary, machine-gradable rubric items used at evaluation time.
The rubric is built and verified in the structured space while grading happens in the converted binary space, retaining the expressiveness of structured rubrics while inheriting the low variance of binary LLM-as-a-judge scoring.

\name (\emph{Grounded Assessment of Multimodal Factuality}) instantiates this design on a setting we call \emph{everyday deep research}: realistic, naturally phrased questions a person would ask a wearable assistant while looking at something, which nonetheless require multi-step research across several web sources.
For instance, a wearer looking at a flaky pastry might ask what technique produces its distinct layers and the science of how they form during baking.
\Cref{fig:teaser} shows one such question and illustrates how \name represents what a complete answer should contain, from the structured meta-rubric to the compiled binary checklist.
This is deep research in the sense of the research process, not academic depth; the questions cover explorative learning, troubleshooting, and shopping, rather than report-level questions on academic topics.
Every question is generated conditioned on a real image drawn from CRAG-MM~\citep{cragmm}, spanning 10 domains that reflect real smart-glasses use, often under imperfect conditions such as low light, blur, or occlusion.
An open-ended, long-form, knowledge-seeking question bank grounded in such imagery is, to our knowledge, itself new.
While \name is multimodal-first, its evaluation framework is modality-agnostic, and we also release a text-only variant whose questions are made self-contained.

Constructing \name involves two tasks, Question Creation and Rubric Creation, each pairing frontier LLMs that can perform web search with expert human annotators over multiple rounds, so that neither the questions nor the reference facts rest on unverified model output.
The resulting benchmark contains \nq questions across 10 domains; \Cref{section:data} describes the process in detail.
Evaluating 14 frontier and open-weight models reveals that \name is genuinely challenging, with the strongest (Gemini 3.1 Pro) reaching a \metric{}\footnote{The \metric{} is an importance-weighted average of how many of a rubric's binary checks a response satisfies, ranging from $-1$ to $1$ with contradictions penalized below omissions (\Cref{subsec:scoring}).} of only \bestmm, highly discriminative, with rankings that align closely with independently understood model capabilities, and robust to the choice of judge.
The text-only variant further shows that removing visual identification raises scores by a roughly constant amount across models, so the multimodal ranking reflects differences in knowledge completeness rather than perception alone.

This work makes four contributions.
First, it makes the case for factual completeness as a first-class, under-studied target, and identifies the reduction of factuality to independent boolean fact-checks as a shared limitation of existing precision and recall methods.
Second, it introduces a two-level rubric representation that reconciles the expressiveness needed for annotation with the reliability needed for automatic grading.
Third, it presents \name, a multimodal-first benchmark of everyday deep-research questions grounded in real wearable imagery, built by combining frontier LLMs with expert human annotators.
Finally, it benchmarks a wide range of frontier models and shows that factual recall is far from solved, with large gaps that precision-based evaluation fails to reveal.

\section{Related Work}
\label{section:related}

\subsection{Long-Form Factuality Evaluation}

Most work on long-form factuality has measured \emph{precision} through a decompose-search-verify (DSV) pipeline: a response is broken into atomic claims, each verified against retrieved web documents \citep{factscore, safe, veriscore, factool, factbench}.
A thinner thread targets \emph{recall}, checking whether a response covers the reference facts a complete answer should contain \citep{verifact, nuggets, factory, olaph}.
Both share a deeper limitation: they represent an answer as a flat list of independent boolean fact-checks, which cannot express the ordering, grouping, and open-ended coverage that determine whether an answer is faithful (\Cref{section:intro}).
A few methods attach a single richer signal to this flat representation, such as event order \citep{dovescore} or importance-weighted recall \citep{importancefact, importancerecall}, but each remains a list of atomic facts with one added axis, far short of representing an answer's content as structured, gradable targets.
This limitation, with DSV's reliance on noisy web search, motivates the structured meta-rubric of \Cref{section:framework}.

\subsection{Rubric-Based Evaluation}

Grading free-form responses with a strong LLM against stated criteria \citep{geval, mtbench} is now a standard alternative to reference-based metrics, but holistic Likert judgments are noisy, which has driven a move to rubric-based grading against an explicit list of criteria \citep{prometheus, flask}, and further to binary checklists, which lower score variance and raise agreement \citep{checkeval, autorubric}.
Applied to factuality and long-form generation, such rubrics grade against expert-written criteria \citep{openscholar, healthbench, researchrubrics, expertlongbench}.
Closest to \name is FACTS Multimodal \citep{factsmm, factsgrounding}, which scores image-grounded answers against a human-authored rubric of reference facts labeled essential or non-essential.
The two differ on both ends: its questions elicit a few essential facts, whereas \name targets richer questions whose complete answers span many facts; and it sorts those facts into a single essential-versus-non-essential split, without the ordering, grouping, importance tiers, or open-ended sets \name represents.
More broadly, prior rubrics are a flat set of criteria authored and graded in the same space.
\name instead authors a per-question, two-level structured meta-rubric and mechanically compiles it into a flat checklist, so annotation happens in the expressive structured space while grading is decomposed into many narrow, self-contained checks.

\subsection{Multimodal Knowledge-Seeking Benchmarks}

\name relates to multimodal retrieval and agentic research benchmarks \citep{gaia, browsecomp, mmsearch, mmbrowsecomp}, which pair each question with a single short, verifiable answer, whereas \name asks open-ended questions and scores factual completeness against a reference-fact rubric.
Closest in setting are multimodal deep-research benchmarks \citep{miroeval, mmdrbench}, which grade the quality of a generated report rather than the factual completeness of an answer.
MMDeepResearch-Bench scores reports with an adaptive LLM-as-a-judge, the holistic scoring \name avoids.
MiroEval instead grades against a rubric, but one of report-level process and outcome criteria rather than the a priori reference facts a complete answer must contain.
We also target \emph{everyday deep research}: realistic wearable-assistant questions that require multi-source research yet stay everyday in topic, rather than the academic or report-level depth those systems aim at.

\section{A Two-Level Meta-Rubric Framework for Evaluating Open-Ended Generation}
\label{section:framework}

A faithful rubric must capture the coverage and ordering among facts that determine a good answer, but a judge asked to weigh that structure while scoring reverts to the holistic, high-variance judgment that rubrics were meant to avoid.
\name resolves this tension with a \textbf{two-level representation}: for each question a structured \emph{meta-rubric} captures how the content of a good answer is organized (\Cref{subsec:metarubric}), which fixed mechanical rules then convert into a flat \emph{binary rubric} of pass-or-fail checks (\Cref{subsec:convert}) that a response is scored against (\Cref{subsec:scoring}).
The expressive structure thus lives where a good answer is described, and the reliable binary checks live where the judge operates.
We develop it for long-form factuality, but it applies to rubric-based evaluation of open-ended generation more broadly.

\subsection{Meta-Rubrics}
\label{subsec:metarubric}

A meta-rubric describes answer quality the way a knowledgeable reviewer reasons about it: which points are essential, which sets of options are acceptable, what must be said in order, and what merely rounds out the answer.
It is a set of \emph{meta-rubric items}, each describing one part of what a complete answer should contain and carrying a \emph{type} that says how that part is structured and an \emph{importance tier}.
The three tiers are \emph{Answer-Critical} for content a correct answer must contain, \emph{Valuable} for content that meaningfully improves it, and \emph{Context} for helpful background.
An item takes one of five types.

\paragraph{Simple Knowledge.} A single discrete fact, such as the identity of the depicted entity or a specific date.

\paragraph{Strict List.} A finite set in which every item is required, such as the core ingredients of a dish.
Naming the set as a list, rather than as separate facts, makes a missing item easier to catch.

\paragraph{Flexible List.} A pool of valid options in which sufficient coverage is required but no specific item is, such as which dishes use a given ingredient.
It specifies a \emph{baseline threshold} of minimum acceptable coverage, for example naming at least two options.
When a concept has both a required core and an optional remainder, it is split into a strict list and a flexible list so each stays internally uniform.

\paragraph{Process.} An \emph{ordered} sequence of steps in which the order carries meaning, such as how an object reached its modern form.
A process may have both required and optional steps.

\paragraph{Relationship.} A connection or contrast between entities, such as how one option compares to an alternative or how one property depends on another.

Beyond the items, a list or process may carry a \emph{meta-insight}: a synthesis across items, such as an overarching trend or categorization, that no single item captures.
For a dish prepared with different red meats, the specific meats form a flexible list, but the pattern that it essentially always uses some red meat is a meta-insight.
Its importance is assigned case by case, since some meta-insights are as essential as the underlying facts while others are merely secondary.

\subsection{From Meta-Rubrics to Binary Rubrics}
\label{subsec:convert}

Asking a judge to weigh coverage, importance, and ordering at once would reintroduce the holistic judgment we set out to avoid, so \name converts each meta-rubric into a flat \emph{binary rubric} of pass-or-fail checks by fixed, mechanical rules, making the conversion deterministic and auditable.
Most of the design lies in how the flexible structures are handled.

\paragraph{Simple knowledge and strict lists.} A simple-knowledge item becomes one check.
A strict list becomes one check per item, each inheriting the list's tier.

\paragraph{Flexible lists.} A flexible list becomes two kinds of check: one or two baseline thresholds and additional credit for broader coverage.
The baseline is a check of the form ``mention at least $N$ of the following: $\dots$'' with all options enumerated so the check is self-contained, inheriting the list's importance.
The additional credit sits one level lower and depends on pool size: a \emph{short} list ($\leq 7$ items) produces one check per item, while a \emph{long} list produces a few higher coverage thresholds instead, since one check per item would flood the rubric with low-value checks.
Since items sit one importance tier below the list but Context has no tier beneath it, a Context-level short list is treated like a long one.
For a short Answer-Critical list of seven dishes, the conversion produces an Answer-Critical check ``mention at least 2 of the following'', enumerating all seven, plus seven Valuable per-dish checks: naming two clears the requirement, naming more raises the Valuable score, and naming a dish outside the seven neither helps nor hurts.
The framework thus measures how much of the acceptable space an answer covers, not whether it matched a fixed list.

\paragraph{Processes.} A process produces one presence check per step, required steps at the list's tier and optional steps one lower.
It adds a sequence check over the required steps at the list's tier, and, when optional steps are present, a second over the full ordering one level lower.

\paragraph{Relationships and meta-insights.} A relationship becomes a check identifying the related entity plus one check per aspect of the connection; each meta-insight becomes a single check at its assigned tier.

Across all types, the checks fall into two groups: those constituting the genuine requirement, which keep their parent item's importance, and secondary ones, namely individual optional items, higher coverage thresholds, optional steps, and the full ordering, which drop one level.
The compiled rubric is again a flat list of binary checks.
But its threshold and sequence checks grade coverage of an open set and correct ordering, which a one-fact-per-check rubric cannot, and each check's tier records how much its content matters.
Appendix~\ref{appendix:examples} works through two complete questions end to end, each showing a full meta-rubric and the binary checklist it compiles into.

\subsection{Scoring}
\label{subsec:scoring}

Given the binary rubric and a response, we compute a \metric{} in three steps: a verdict on each check, a score within each tier, and a weighted combination across tiers.

\paragraph{Per-check verdicts.} An LLM judge evaluates the response against each check independently and returns one of four verdicts.
A check can fail in two ways that a binary met-or-not verdict would collapse: the response can \emph{miss} the required content or \emph{contradict} it.
Scoring these the same would treat a cautious answer and a confidently wrong one alike, so we separate them and penalize contradiction more heavily, which is essential for a benchmark whose purpose is factual reliability.
We further distinguish \emph{meets} from \emph{partially meets} to credit a response that is directionally right but imprecise, such as the correct genus but not species.

\paragraph{Per-tier score.} Verdicts map to numbers: \emph{meets} scores $1$, \emph{partially meets} scores $\lambda$, \emph{missing} scores $0$, and \emph{contradicts} scores $\mu$, with $0 < \lambda < 1$ and $\mu < 0$ so a falsehood costs more than an omission.
Writing $M$, $P$, $S$, and $C$ for the counts of each verdict, the tier score is
\begin{equation}
  s = \frac{M + \lambda\,P + \mu\, C}{M + P + S + |\mu|\, C},
\end{equation}
which lies in $[-1, 1]$.
Because $\mu$ enters both numerator and denominator, a contradiction both removes credit and counts as more than one check, so a cautious answer outranks a confident but wrong one.
We use $\lambda = 0.5$ and $\mu = -2$.

\paragraph{Combining tiers.} The \metric{} is a weighted average of the three tier scores with global weights $w_{\text{AC}}$, $w_{\text{V}}$, $w_{\text{C}}$, set so that covering the Answer-Critical content alone reaches roughly a passing score.
The weights are global rather than implied by per-tier check counts, which vary widely across questions, so that a question with many minor Context checks cannot drown out its few Answer-Critical ones.
We use $w_{\text{AC}} = 0.6$, $w_{\text{V}} = 0.3$, $w_{\text{C}} = 0.1$, and when a question has no checks in a tier its weight is redistributed over the rest.

\section{Dataset Construction}
\label{section:data}

\name is built in two stages: we create \emph{everyday deep research} questions grounded in real wearable imagery (\Cref{subsec:qcreation}), then construct a structured meta-rubric for each and convert it into the binary rubric used for scoring (\Cref{subsec:rubriccreation}).
Both stages pair a frontier LLM, which proposes candidates at scale, with human annotators who review and revise them over multiple rounds, so the final questions and rubrics rest on verified human judgment rather than unchecked model output.

\subsection{Question Creation}
\label{subsec:qcreation}

The goal is to elicit questions a person would naturally ask a wearable assistant about something in view, yet that require genuine multi-step research to answer.
The central difficulty is that such a question must depend on the image without describing the entity in words: if it names or characterizes its subject, it is answerable from text alone and the image is decorative; if it presupposes facts only someone who already recognized the entity would know, it is leading rather than information-seeking.
A good question therefore refers to its subject the way a user would, through a pronoun or generic term such as ``this car'' or ``these berries'', and still admits a single research-intensive answer.

\paragraph{Source images.}
We draw 1{,}938 images and their ground-truth entities from CRAG-MM~\citep{cragmm}, a benchmark of wearable-device imagery.
Most images (roughly 80\%) are egocentric photographs taken with smart glasses, in which the object of interest is often small, rotated, truncated, occluded, or poorly lit.
The entities span 10 domains of everyday life, from plants, food, and animals to vehicles, local places, everyday objects, and consumer products, and range across head, torso, and tail popularity.

\paragraph{Candidate generation.}
For each image we prompt a frontier multimodal LLM\footnote{We use Gemini~3.1~Pro throughout question and rubric construction.} to propose open-ended candidate questions (Appendix~\ref{appendix:prompt:qgen}) that require multi-step web research and a multi-paragraph answer, are concise and natural as a user would actually speak, and refer to the entity only through image-dependent terms.

\paragraph{Multi-round human revision.}
Annotators then \emph{accept}, \emph{revise}, or \emph{discard} each candidate, with every question reviewed independently by two annotators and a third adjudicating disagreements.
A question accepted by at least two annotators advances, a revised one re-enters the next round, and when an image's candidates are all discarded, new ones are generated.
The process repeats until every image has a question that passes human review, and only a question accepted by both the annotators and the automatic pass below enters \name.

\paragraph{Automatic filtering, ranking, and diversification.}
A second LLM pass (Appendix~\ref{appendix:prompt:qselect}) corrects two recurring errors of the initial per-image generation.
First, many candidates are \emph{leading}: they presuppose the entity is already recognized, making the image decorative.
It rejects any question failing a hard requirement, most importantly a \emph{stranger test}: a question is rejected as leading if it could only be asked by someone who already knew the entity's identity.
For example, ``how does this engine's turbocharger improve on the previous generation?'' presumes the car and its engine history are already recognized, whereas ``what kind of car is this and how does it perform?'' is something a stranger could ask from the image alone.
Second, because questions are generated independently per image, they tend to cluster into a few templates.
Among passing questions it selects the one that is at once concise, research-intensive, and specific to the entity rather than boilerplate, and it proposes additional diverse candidates that are seeded into annotation for images whose only accepted questions are boilerplate.
This loop continues until every image has a question accepted by both the annotators and the LLM, which becomes the image's entry in \name.
The procedure yields 1{,}843 questions, one per image; we analyze their diversity in \Cref{section:analysis}.

\subsection{Rubric Creation}
\label{subsec:rubriccreation}

For each question we construct the structured meta-rubric and the binary rubric it converts to, following \Cref{section:framework}.
As with questions, an LLM produces the initial rubric at scale and expert annotators verify and revise it, but rubric creation is far more demanding: it requires gathering the facts a complete answer should contain, organizing them into the right structures, and checking the conversion into the binary rubric used for evaluation.

\paragraph{Generation.}
We prompt the LLM (Appendix~\ref{appendix:prompt:rubricgen}) to first research the question on the web, then build a structured meta-rubric from the retrieved knowledge, citing the supporting \emph{web snippets}, and compile it into the binary rubric by the rules of \Cref{section:framework}.

\paragraph{Self-refinement.}
Because a one-pass draft can be incomplete or imperfectly grounded, the LLM then refines its own rubric (Appendix~\ref{appendix:prompt:rubricrefine}): an audit phase re-verifies every snippet by browsing its source and flags partially covered lists or processes, and an independent search phase re-researches the question from scratch to retrieve the missing evidence rather than editing the draft in place.
Refinement operates on the structured meta-rubric, and a final phase rebuilds it as a verified superset of the draft and re-compiles it into the binary rubric by the rules of \Cref{section:framework}.
A second refinement round changed little, so we use one.

\paragraph{Multi-round human revision.}
The refined rubrics are revised by a small in-house team working closely with the authors, since a general annotation pool could not improve on the LLM-generated rubrics, which demand both subject-matter judgment and fluency with the structured representation.
Annotators are shown the meta-rubric but revise the binary rubric directly, because it is what is used for scoring, so revising it lets them check the conversion itself; they are trained to reason in the structured space, using the meta-rubric as a scaffold to spot a missing list item, a misordered step, or a miscategorized fact.
After a first revision round we run another pass of LLM refinement followed by a second human round.

\paragraph{Evidence snippets.}
To make the rubrics more verifiable and grounded against web sources, we also provide \emph{evidence snippets} which the rubrics can cite to support its claims.
During generation the LLM searches the web and attaches the supporting snippets to each rubric element it writes.
The annotators then refine these snippets alongside the rubric, browsing each cited source, dropping ones that do not support their element, and adding evidence the model missed.

\paragraph{Entity and question corrections.}
Working through the rubrics surfaced a few upstream errors, cases where the CRAG-MM ground-truth entity was wrong or a question was ambiguous about which object it referred to, which we corrected by hand in 12 examples.
A small number of questions were dropped when no sound rubric could be produced, leaving \nq questions with complete, human-verified rubrics.
Annotator training and auditing details for both stages are given in Appendix~\ref{appendix:annotation}.

\begin{table*}[t]
  \centering
  \caption{Model performance on the \nq multimodal questions of \name, judged by \judge and scored with the \Cref{subsec:scoring} procedure. \metric{} is the headline weighted score (range $[-1, 1]$); the per-tier scores (AC / Val / Ctx) are the unweighted scores within each importance tier; both are shown as percentages. The verdict distribution gives the percentage of rubric elements receiving each verdict.}
  \label{table:main}
  \begin{tabular}{llcccccccccc}
    \toprule
    & & & & \multicolumn{3}{c}{Per-tier score} & & \multicolumn{4}{c}{Verdict distribution (\%)} \\
    \cmidrule(lr){5-7} \cmidrule(lr){9-12}
    & Model & \name & & AC & Val & Ctx & & meets & partial & missing & contra. \\
    \midrule
    \multirow{8}{*}{\rotatebox{90}{Proprietary}}
    & Gemini~3.1~Pro           & \textbf{58.7} & & 63.9 & 51.9 & 45.1 & & 55.2 & 11.5 & 29.2 & 4.1 \\
    & Gemini~3~Flash           & 57.9 & & 63.0 & 51.6 & 44.7 & & 54.9 & 12.1 & 28.5 & 4.5 \\
    & Claude~Opus~4.8          & 53.1 & & 59.4 & 43.9 & 40.4 & & 45.3 & 13.5 & 38.6 & 2.6 \\
    & Gemini~2.5~Pro           & 51.8 & & 55.8 & 47.2 & 39.7 & & 52.1 & 11.5 & 30.5 & 5.9 \\
    & Claude~Opus~4.6          & 50.5 & & 54.6 & 45.7 & 38.8 & & 49.2 & 12.9 & 32.6 & 5.2 \\
    & GPT-5.4                  & 42.6 & & 46.8 & 37.8 & 29.0 & & 39.5 & 14.0 & 42.0 & 4.5 \\
    & Claude~Sonnet~4.6        & 41.5 & & 46.7 & 34.3 & 30.5 & & 38.4 & 13.5 & 43.2 & 4.9 \\
    & GPT-4o                   & 34.2 & & 39.1 & 28.0 & 21.5 & & 29.7 & 12.6 & 54.3 & 3.4 \\
    \midrule
    \multirow{6}{*}{\rotatebox{90}{Open-weight}}
    & Qwen3-VL~235B            & 33.0 & & 34.1 & 32.6 & 26.4 & & 39.4 & 12.1 & 39.8 & 8.7 \\
    & Llama~4~Maverick         & 18.3 & & 19.5 & 17.3 & 12.3 & & 22.9 & 11.7 & 57.9 & 7.6 \\
    & Llama~4~Scout            & 14.1 & & 14.4 & 14.3 & 10.5 & & 21.1 & 11.3 & 59.3 & 8.3 \\
    & Qwen3-VL~8B              & 13.7 & & 12.3 & 17.0 & 11.5 & & 27.3 & 10.6 & 50.7 & 11.5 \\
    & Llama~3.2~90B            & 11.8 & & 12.7 & 10.9 & 8.2 & & 17.8 & 10.0 & 65.1 & 7.2 \\
    & Llama~3.2~11B            & 5.1 & & 3.8 & 7.8 & 4.3 & & 14.2 & 9.2 & 68.3 & 8.3 \\
    \bottomrule
  \end{tabular}
\end{table*}

\section{Evaluating Models on \name}
\label{section:eval}

We score every model with the procedure of \Cref{subsec:scoring}, using \judge as the LLM judge (Appendix~\ref{appendix:prompt:judge}) and weights $w_{\text{AC}}=0.6$, $w_{\text{V}}=0.3$, $w_{\text{C}}=0.1$ with $\lambda=0.5$, $\mu=-2$.
Each model is given the image and question and produces a free-form answer, which the judge grades against the binary rubric.
Two additional judges reproduce the same ranking, and no judge favors its own answers (\Cref{table:judges}), so we report \judge scores throughout.

\begin{table}[ht]
  \centering
  \caption{Judge agreement: all 14 models scored by three different judges on all \nq questions. \judge and Claude Opus 4.8 give an identical ranking; the weaker Qwen3-VL 235B is more lenient and preserves the same broad ordering. Scores are \metric{} percentages.}
  \label{table:judges}
  \begin{tabular}{lccc}
    \toprule
    Model & Gemini & Claude & Qwen3-VL 235B \\
    \midrule
    Gemini~3.1~Pro       & 58.7 & 58.7 & 67.0 \\
    Gemini~3~Flash       & 57.9 & 57.4 & 66.1 \\
    Claude~Opus~4.8      & 53.1 & 52.8 & 60.1 \\
    Gemini~2.5~Pro       & 51.8 & 51.7 & 60.9 \\
    Claude~Opus~4.6      & 50.5 & 50.1 & 59.1 \\
    GPT-5.4              & 42.6 & 41.3 & 51.0 \\
    Claude~Sonnet~4.6    & 41.5 & 40.4 & 48.3 \\
    GPT-4o               & 34.2 & 33.6 & 38.8 \\
    Qwen3-VL~235B        & 33.0 & 31.2 & 44.0 \\
    Llama~4~Maverick     & 18.3 & 16.7 & 22.1 \\
    Llama~4~Scout        & 14.1 & 12.8 & 18.6 \\
    Qwen3-VL~8B          & 13.7 & 12.6 & 23.0 \\
    Llama~3.2~90B        & 11.8 & 10.1 & 16.3 \\
    Llama~3.2~11B        & 5.1 & 3.3 & 9.6 \\
    \bottomrule
  \end{tabular}
\end{table}

\subsection{Main Results}
\label{subsec:mainresults}

\Cref{table:main} reports the \metric{}, per-tier scores, and verdict distribution for each model on the \nq multimodal questions.
Three findings stand out.
First, \name is challenging: the best model reaches a \metric{} of only \bestmm and meets at most \bestmeets of the rubric elements, leaving substantial headroom.
Second, the metric is highly discriminative, and its ranking recovers regularities independently understood about these models: newer and larger models outscore older and smaller ones, proprietary outscore open-weight, and known tendencies surface in the verdicts, such as the two Qwen3-VL variants showing the highest contradiction rates and Claude Opus the lowest at 3\%.
Reproducing these orderings without being tuned to them is evidence the \metric{} captures real answer quality rather than noise.
Third, the dominant failure is omission rather than error: \emph{missing} verdicts account for roughly two-thirds of the rubric elements for the weakest models and over a quarter even for the strongest, so what separates models is largely how much of a complete answer they supply, while contradiction rates stay low for the strongest and climb sharply for the weaker ones.

\paragraph{Judge agreement.}
To test whether the rankings depend on the choice of judge, we re-score all 14 models with two additional judges, Claude Opus 4.8 and Qwen3-VL 235B, on all \nq questions with the same rubrics and scoring procedure (\Cref{table:judges}).
Gemini and Claude produce an identical ranking, with per-model scores within 1.8 points of each other, and neither favors its own answers: each scores itself within 0.3 points of how the other scores it, so \judge does not inflate its own answers.
Qwen3-VL 235B, a smaller open-weight judge, is uniformly more lenient by 4 to 11 points but preserves the same broad ordering, with only adjacent swaps among closely spaced models.
The \metric{} is thus robust to the judge, and the rankings in \Cref{table:main} are not artifacts of judge-specific biases.

\begin{table*}[ht]
  \centering
  \caption{Multimodal versus text-only \metric{} on the same 1{,}806 questions.}
  \label{table:textonly}
  \begin{tabular}{lccc}
    \toprule
    Model & Multimodal & Text-only & $\Delta$ \\
    \midrule
    Gemini~3.1~Pro            & 58.7 & 73.7 & +15.0 \\
    Gemini~3~Flash            & 57.9 & 70.8 & +12.9 \\
    Claude~Opus~4.8           & 53.1 & 59.9 & +6.8 \\
    Gemini~2.5~Pro            & 51.8 & 66.2 & +14.4 \\
    Claude~Opus~4.6           & 50.5 & 62.1 & +11.6 \\
    GPT-5.4                   & 42.6 & 58.5 & +16.0 \\
    Claude~Sonnet~4.6         & 41.5 & 53.8 & +12.2 \\
    GPT-4o                    & 34.2 & 56.5 & +22.3 \\
    Qwen3-VL~235B             & 33.0 & 47.2 & +14.2 \\
    Llama~4~Maverick          & 18.3 & 36.5 & +18.2 \\
    Llama~3.2~90B             & 11.8 & 25.2 & +13.4 \\
    Llama~4~Scout             & 14.1 & 27.8 & +13.7 \\
    Qwen3-VL~8B               & 13.7 & 23.4 & +9.7 \\
    Llama~3.2~11B             & 5.1 & 17.7 & +12.7 \\
    \bottomrule
  \end{tabular}
\end{table*}

\subsection{A Text-Only Variant}
\label{subsec:textonly}

Because our meta-rubric framework does not depend on the image, the same rubrics can grade a purely textual task.
We construct and release a \emph{text-only} variant in which each question is rewritten to name its subject explicitly (Appendix~\ref{appendix:prompt:textonly}), keeping the judge and scoring unchanged.
The same conversion step also removes the Answer-Critical rubric item that identifies the named subject, which is trivially satisfied once the question states the subject.
Of the \nq questions, 7 remain image-dependent even after conversion and are excluded, leaving 1{,}806 answerable from text alone.

\paragraph{Results.}
\Cref{table:textonly} reports text-only scores alongside the multimodal ones.
Every model scores higher without the image and the ordering is broadly preserved, so the two settings measure the same underlying quality, and their difference isolates the cost of visual identification.
That cost is rather uniform: removing the image raises the \metric{} by roughly 10 to 20 points for nearly every model, largely independent of its multimodal score.
Visual identification is therefore closer to a constant tax than the axis separating strong from weak models; even the strongest model gains as much as most others, so its remaining gap in the text-only setting reflects genuine incompleteness rather than perception.
Because the gain is broadly constant, the multimodal ranking survives the removal of the image.

\section{Dataset Analysis}
\label{section:analysis}

We analyze both the questions and the rubrics of \name.
The questions are diverse and complex while staying everyday, the \emph{everyday deep research} setting the benchmark targets (\Cref{subsec:qdiversity}).
The rubrics serve two purposes in our analysis.
First, they showcase the meta-rubric structure at the core of our framework (\Cref{subsec:metastats}).
We show this structure is real, appears in almost every question, and is useful, since a flat rubric would misrepresent most questions.
Second, the depth of a rubric reflects how demanding its question is, since a rubric enumerates what a complete answer must contain (\Cref{subsec:rubricstats}).

\subsection{Question Diversity}
\label{subsec:qdiversity}

\begin{figure*}[ht]
  \centering
  \includegraphics[width=\textwidth]{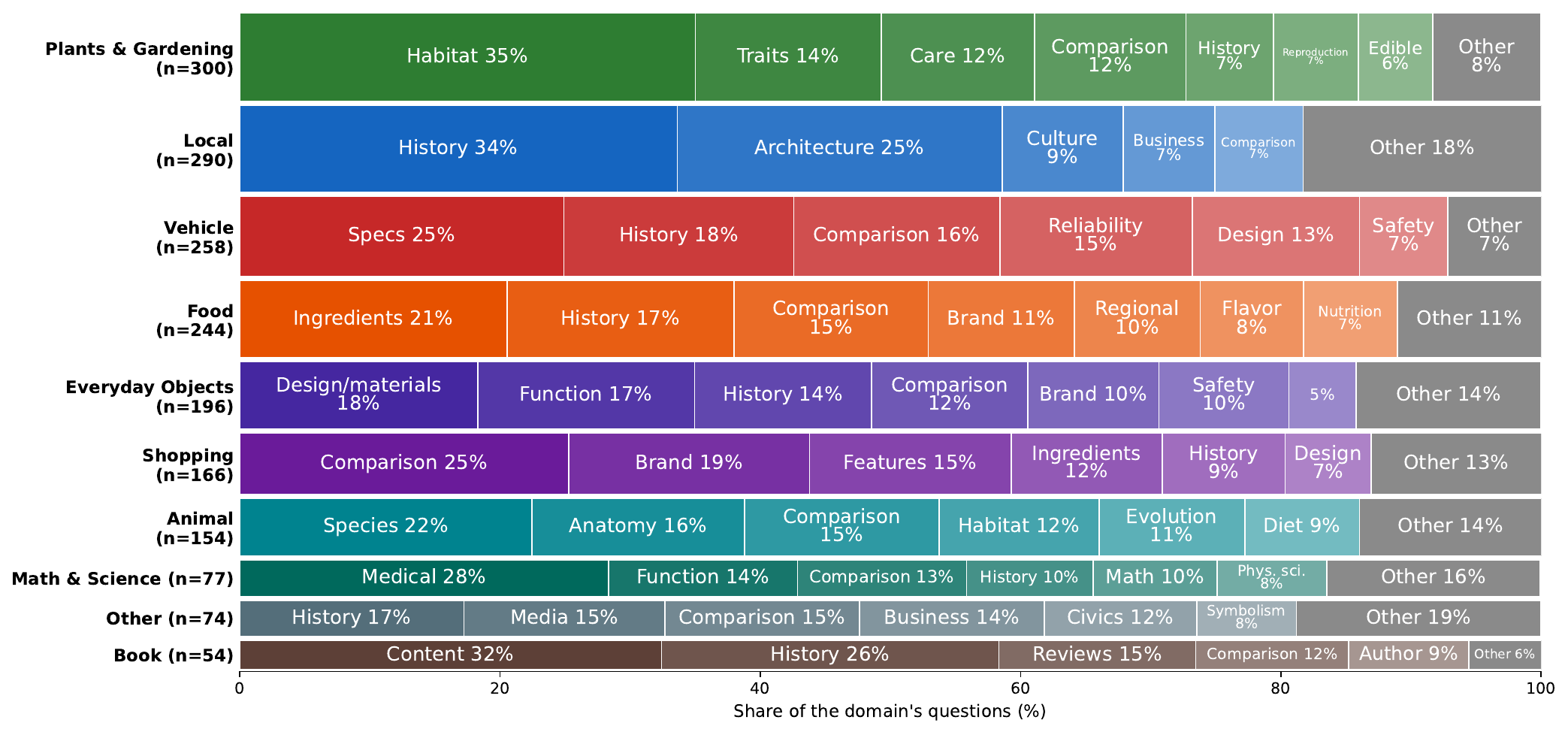}
  \caption{Topic composition of \name questions within each of the 10 domains. Row height is proportional to the number of questions in the domain; horizontal segments give the share touching each topic, ranked by frequency, with shares below 5\% folded into \emph{Other}. The distinct per-domain profiles show questions are tailored to their subject rather than drawn from a few shared templates.}
  \label{fig:domains}
\end{figure*}

Image-grounded question sets risk collapsing into a few templates, such as ``what is this and what is it used for'' repeated across every image, which the pipeline of \Cref{subsec:qcreation} counters in both surface form and content.
On the surface the questions are short (median 23 words) yet not templated: the 25 most common opening phrases cover under 60\% of them, and about one in six do not begin with a standard interrogative, opening instead with framings such as ``Based on \dots'' or ``If I wanted \dots''.
In content each domain spans several recurring topics rather than one dominant template (\Cref{fig:domains}): questions about plants range over habitat and care, those about local places over history and architecture, those about vehicles over specifications and reliability, and those about food over ingredients and regional variation.
The topics a domain covers are naturally characteristic of its subject, and the pipeline spreads questions across them rather than letting a single topic dominate.

\begin{figure*}[ht]
  \centering
  \includegraphics[width=\textwidth]{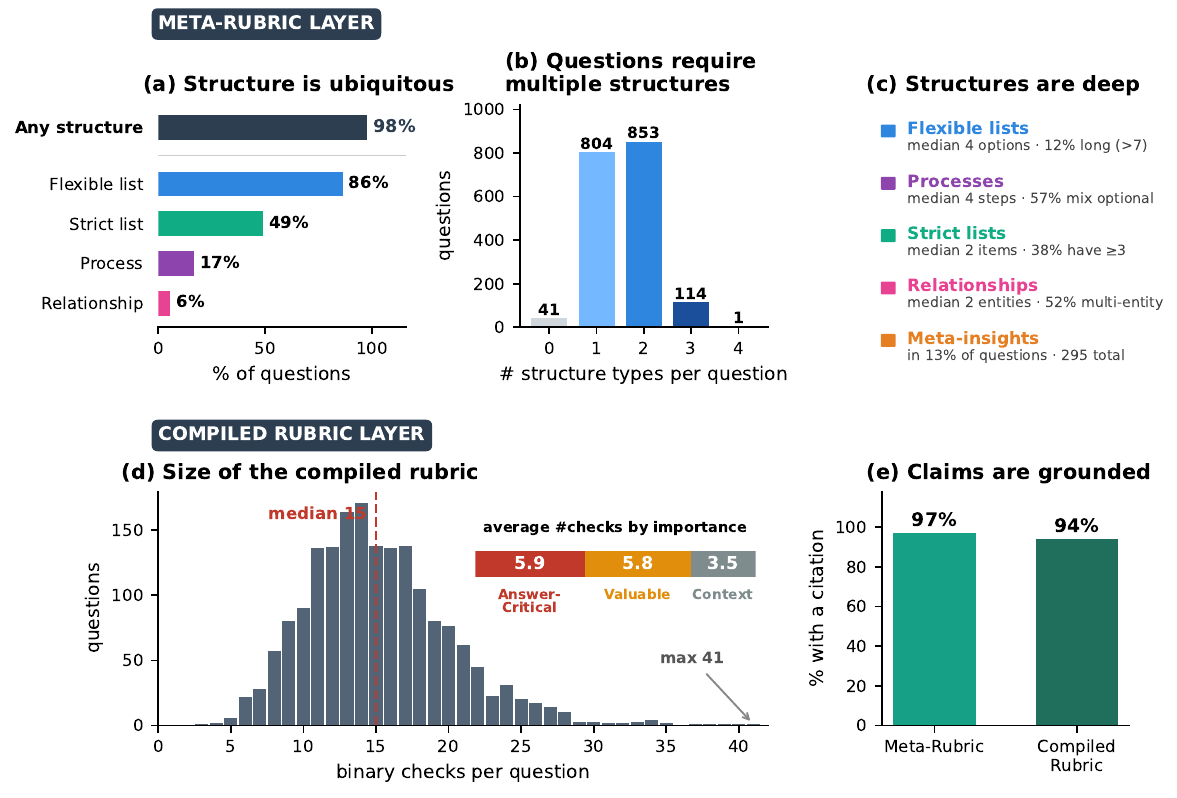}
  \caption{Rubric statistics for \name ($n=\nq$). The structure the two-level design captures is pervasive, not incidental: almost every question needs at least one component beyond plain facts, most need several, and those components are deep, so a flat checklist would misrepresent the large majority of questions. The depth of the rubrics also reflects how demanding the questions are, since each rubric enumerates what a complete answer requires. Panels: (a) prevalence of each meta-rubric type, (b) how many types co-occur per question, (c) depth of the structures, (d) size and importance-tier breakdown of the compiled checks, (e) evidence grounding at both levels: 97\% of meta-rubric items and 94\% of compiled checks cite a supporting web snippet.}
  \label{fig:rubricstats}
\end{figure*}

\subsection{Rubric Statistics}
\label{subsec:rubricstats}

Each question carries a rubric of \avgchecks binary checks on average (median 15), split across the three tiers at roughly \accheck Answer-Critical, \valcheck Valuable, and \ctxcheck Context checks (\Cref{fig:rubricstats}).
Every check is grounded in human-verified evidence: a rubric draws on about 10 web snippets, and across the benchmark these cite over 9{,}400 distinct web pages, so the rubrics rest on a broad base of external references.
This scale, roughly 15 independently checkable, evidence-backed criteria per question, is what lets the \metric{} measure completeness at a fine grain rather than as a single holistic judgment.
The Answer-Critical and Valuable tiers carry comparable numbers of checks (\accheck vs \valcheck), and the two lower tiers together, Valuable and Context, account for most of a rubric: beyond the strictly required core, a complete answer must cover an open-ended space of facts that meaningfully improve it, and it is this middle ground where models differ most.

\subsection{The Structure Behind the Rubrics}
\label{subsec:metastats}

The counts above describe each rubric as a flat list of independent checks, but that picture is incomplete: the checks are compiled from a structured meta-rubric, and the structure is pervasive.
Only a small fraction of questions are answerable by isolated facts alone: 98\% have a rubric with at least one structured component beyond Simple Knowledge, with more than half requiring two or more structures.
Flexible lists are the most common (86\% of questions), reflecting how often a good answer must cover enough of an open-ended set; 49\% also contain a strict list of jointly required items, 17\% a process whose steps must be ordered, and 6\% an explicit relationship.
Each structure materializes as a check a conventional flat rubric would not contain: flexible lists produce threshold checks, and processes produce sequence checks that grade ordering.

The structures are also deep rather than nominal.
Flexible-list pools range from a few options to several dozen (median four), and 357 are long enough that tiered coverage thresholds replace per-item checks; processes carry a median of four steps, and 182 of 318 mix optional with mandatory steps, so the ordering and partial-credit logic of \Cref{section:framework} are exercised rather than hypothetical.
Beyond item coverage, 13\% of questions carry a meta-insight, an overarching pattern or end-state a complete answer should convey, which a per-fact checklist has no place to record.

This is the empirical case for the two-level design.
For 98\% of \name's questions a flat rubric of independent, individually required checks would be wrong: it would force a specific item from an open-ended pool or drop the ordering of a process.
The meta-rubric is what lets \name state, per question, which facts are jointly required, which open sets need coverage, what must be ordered, and what merely helps, then compile all of it into checks a judge can grade reliably.

\section{Conclusion}
\label{section:conclusion}

We introduced \name, a benchmark for the factual completeness of long-form, open-ended generation built on a two-level rubric framework.
The framework separates the representation used to describe a good answer from the one used to grade it: a structured meta-rubric captures the open-ended sets, ordered processes, relationships, and importance tiers a faithful answer involves, which is mechanically converted into a flat binary rubric an LLM judge scores reliably.
This resolves the tension between expressiveness and reliability in rubric-based evaluation, and our analysis shows the structure is not incidental: 98\% of \name's questions require a component beyond isolated facts that a flat checklist cannot represent.

\name instantiates the framework on \nq everyday deep-research questions, each grounded in a real image from wearable and public sources and paired with an evidence-backed rubric produced by frontier LLMs and verified by expert annotators.
Evaluating a range of models shows the benchmark is far from solved, with the strongest (Gemini 3.1 Pro) reaching a \metric{} of only \bestmm, and highly discriminative, separating current systems with stable rankings and large margins that hold across different LLM judges.
Because the framework is modality-agnostic, a text-only variant isolates perception from knowledge, showing that naming the subject helps every model by a roughly constant amount and that the overall ranking largely reflects differences in knowledge completeness.

While we develop it for long-form factuality, the two-level meta-rubric is a general recipe for rubric-based evaluation of open-ended generation where a flat list of binary criteria falls short.
We release \name, its text-only variant, and our evaluation scripts.

\section*{Statement on the Use of Large Language Models}
Large language models are used throughout this project, in every case under human review.
Gemini 3.1 Pro proposes candidate questions and rubrics during dataset construction (\Cref{section:data}), which expert annotators then verify and revise, so the released data rests on human judgment rather than unchecked model output.
Claude Opus 4.8 assists with the implementation, the drafting and revision of this manuscript, and the figures and charts.
Both models are also used for brainstorming and discussion.
The authors take full responsibility for all content.

\clearpage
\newpage
\bibliographystyle{assets/plainnat}
\bibliography{paper}

\clearpage
\newpage
\beginappendix

\section{Prompts}
\label{appendix:prompts}

This appendix collects the full text of the prompts used to construct \name.

\subsection{Question Generation}
\label{appendix:prompt:qgen}
The prompt used to generate candidate questions for each image (\Cref{subsec:qcreation}).
\promptbox{Question Generation Prompt}{prompts/question_generation.txt}

\subsection{Question Selection}
\label{appendix:prompt:qselect}
The prompt used to filter, rank, and diversify the accepted questions (\Cref{subsec:qcreation}).
\promptbox{Question Selection Prompt}{prompts/question_selection.txt}

\subsection{Rubric Generation}
\label{appendix:prompt:rubricgen}
The prompt used to generate the initial meta-rubric and binary rubric for each question (\Cref{subsec:rubriccreation}).
\promptbox{Rubric Generation Prompt}{prompts/rubric_generation.txt}

\subsection{Rubric Refinement}
\label{appendix:prompt:rubricrefine}
The prompt used for LLM self-refinement of the draft rubric (\Cref{subsec:rubriccreation}).
\promptbox{Rubric Refinement Prompt}{prompts/rubric_refinement.txt}

\subsection{LLM Judge}
\label{appendix:prompt:judge}
The system prompt and evaluation template used by the LLM judge to grade each binary rubric check (\Cref{subsec:scoring}).
\promptbox{LLM Judge Prompt}{prompts/judge.txt}

\subsection{Text-Only Question Conversion}
\label{appendix:prompt:textonly}
The prompt used to convert each multimodal question into a self-contained text-only question by resolving the entity from the rubrics, and to remove the identification rubric that the converted question makes trivial (\Cref{subsec:textonly}).
\promptbox{Text-Only Conversion Prompt}{prompts/text_only_conversion.txt}

\section{Full Examples}
\label{appendix:examples}

\Cref{fig:teaser} shows a partial example in the main text.
Here we give two complete examples, reproduced verbatim from \name, that together exercise every meta-rubric type and every conversion rule of \Cref{section:framework}.
Each figure shows the question and its human-verified web evidence, the full structured meta-rubric (\Cref{subsec:metarubric}), and the complete binary checklist it compiles into (\Cref{subsec:convert}), with each check placed in its importance tier.

\begin{figure*}[p]
  \centering
  \includegraphics[width=\textwidth,height=0.92\textheight,keepaspectratio]{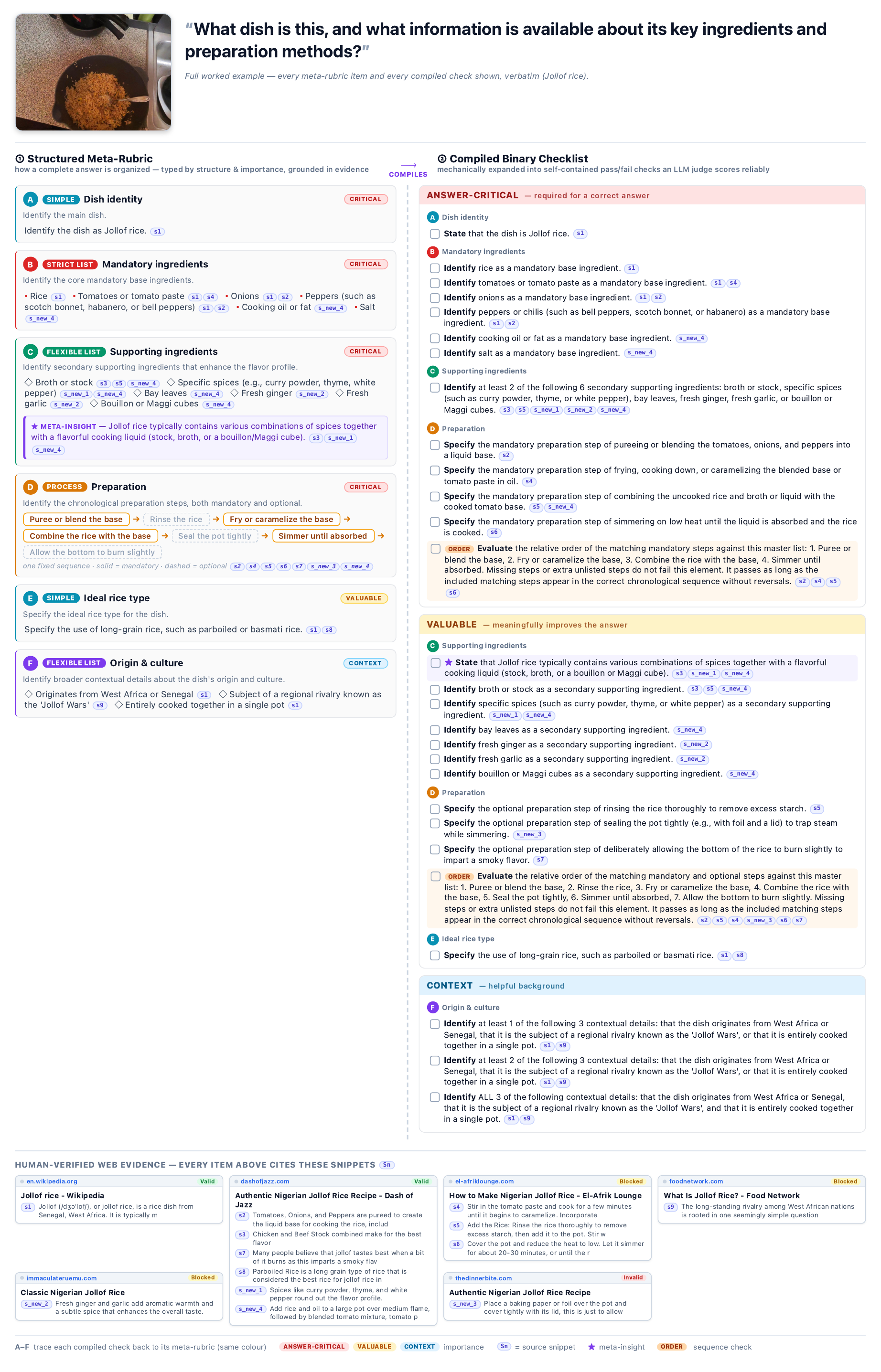}
  \caption{A complete example from \name (jollof rice), the full version of \Cref{fig:teaser}. Left: the structured meta-rubric. Right: the compiled binary checklist, grouped by importance tier. Bottom: the human-verified web evidence, grouped by source page. All text is reproduced verbatim.}
  \label{fig:example-jollof}
\end{figure*}

\Cref{fig:example-jollof} is the complete version of the jollof rice example from \Cref{fig:teaser}.
Its Flexible List of supporting ingredients is Answer-Critical, so its baseline threshold check stays Answer-Critical while its per-item checks drop to Valuable, the one-level demotion of \Cref{subsec:convert}.
Its Process has both mandatory and optional steps, so it produces one sequence check over the mandatory steps at Answer-Critical and a second over the full ordering at Valuable.
The meta-insight attached to the Flexible List becomes a single Valuable check.

\begin{figure*}[p]
  \centering
  \includegraphics[width=\textwidth,height=0.92\textheight,keepaspectratio]{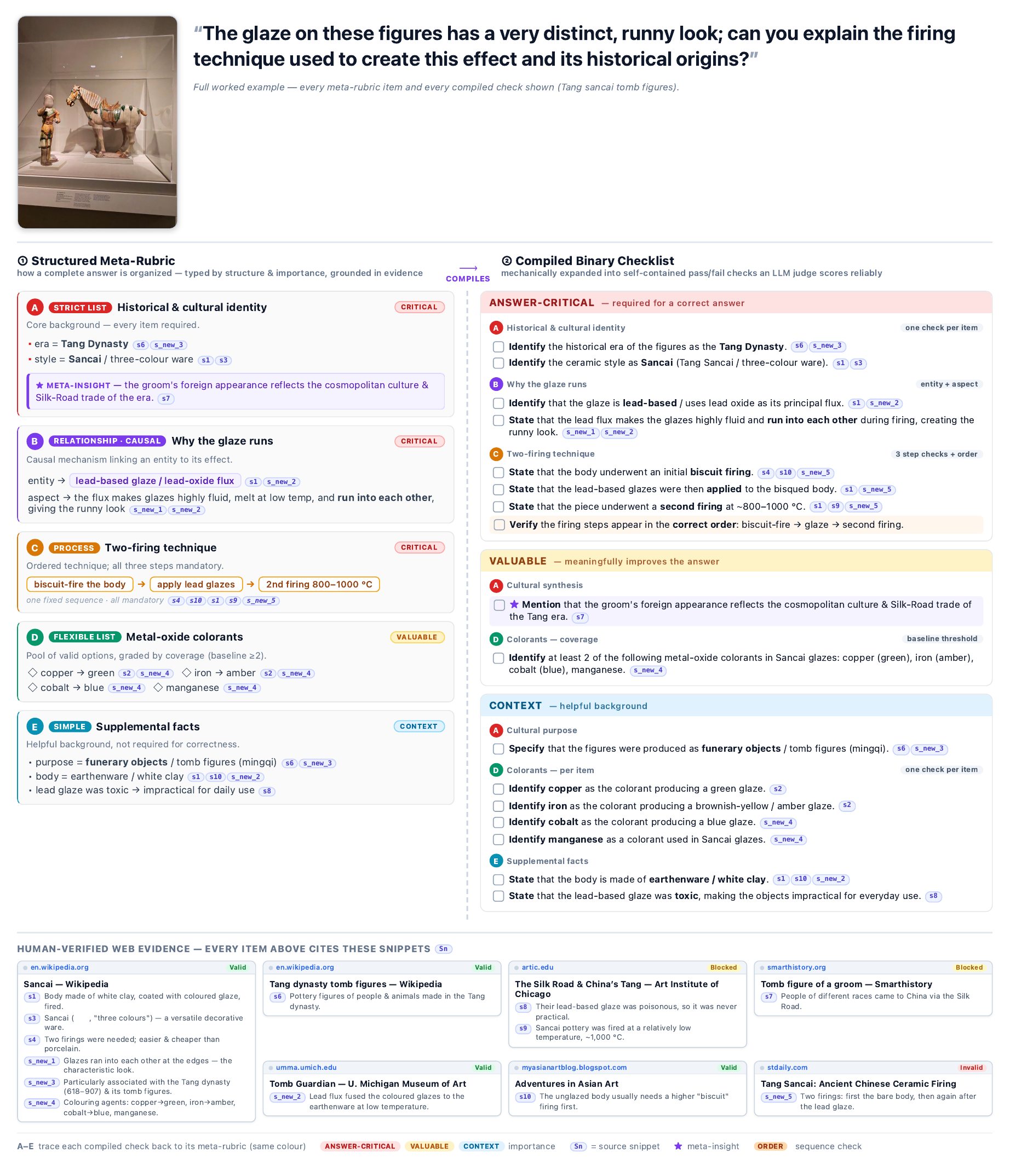}
  \caption{A complete example from \name (a Tang \emph{sancai} figure), exercising all five meta-rubric types including a Relationship. Layout as in \Cref{fig:example-jollof}. All text is reproduced verbatim.}
  \label{fig:example-glaze}
\end{figure*}

\Cref{fig:example-glaze} is a Tang \emph{sancai} figure, and it exercises the one type the jollof example does not: a Relationship, connecting the lead-based glaze to why it runs during firing.
It also places a meta-insight inside a Strict List rather than a Flexible List, and it uses all five meta-rubric types in a single question.
Together the two examples cover Simple Knowledge, Strict List, Flexible List, Process, Relationship, and meta-insights, and show how each compiles into binary checks whose importance tier follows the rules of \Cref{subsec:convert}.

\section{Annotator Training and Auditing}
\label{appendix:annotation}

\paragraph{Question creation.}
Annotators were given guidelines describing the task, the criteria for a good question (naturalness, requires multi-step search, image-dependent, and so on), and worked examples.
They then completed scored gold jobs, and those passing at 95\% moved to the live task.
Completed jobs were audited at 10\% and accepted or rejected with feedback, and further rounds followed the two-way review with third-annotator adjudication described in \Cref{subsec:qcreation}.

\paragraph{Rubric creation.}
Annotators watched a task-and-tool video, then took a guidelines quiz (10 two-part items, each a true/false paired with a related multiple-choice question); those passing at 90\% advanced.
After practice examples, they attended two 45-minute trainings, one on meta-rubrics and one on binary rubrics, each with a Q\&A session that prompted further clarifications to the guidelines, and a chat channel connected annotators with the authors throughout.
Completed jobs were audited at 10\% and returned with comments and edge-case feedback for revision.

\end{document}